\documentclass{article} 
\usepackage[utf8]{inputenc}
\usepackage{iclr2026_conference,times}
\usepackage{enumitem}

\usepackage{amsmath,amsfonts,bm}

\def\eqref#1{equation~\ref{#1}}

\def\1{\bm{1}}

\DeclareMathAlphabet{\mathsfit}{\encodingdefault}{\sfdefault}{m}{sl}
\SetMathAlphabet{\mathsfit}{bold}{\encodingdefault}{\sfdefault}{bx}{n}

\usepackage{booktabs}
\usepackage{hyperref}
\usepackage{url}
\usepackage[ruled,vlined,linesnumbered]{algorithm2e}
\usepackage{amsmath,amssymb}
\usepackage[utf8]{inputenc}
\usepackage{textgreek}
\usepackage{textcomp}
\usepackage{tikz}
\usetikzlibrary{positioning,arrows.meta}
\usepackage{array}
\usepackage{subcaption}  
\usepackage{listings}
\lstdefinelanguage{json}{
  morestring=[b]",
  morecomment=[l]{//},
  morecomment=[s]{/*}{*/},
  morekeywords={true,false,null},
  sensitive=false,
  alsoletter={:},
  literate={
    {"}{{\texttt{\char34}}}1
    {,}{{\texttt{,}}}1
    {:}{{\texttt{:}}}1
    {[}{{\texttt{[}}}1
    {]}{{\texttt{]}}}1
    {{\{}}{{\texttt{\{}}}1
    {{\}}}{{\texttt{\}}}}1
  },
}
\usepackage{xcolor}

\title{Two-Stage Voting for Robust and Efficient Suicide Risk Detection on Social Media}

\author{
\begin{tabular}[t]{ll}
Yukai Song & Pengfei Zhou \\
Department of Electrical and Computer Engineering & Department of Informatics and Networked Systems \\
University of Pittsburgh & University of Pittsburgh \\
Pittsburgh, PA, USA & Pittsburgh, PA, USA \\
\texttt{yukai.song@pitt.edu} & \texttt{pengfeizhou@pitt.edu} \\
\\[-3pt]
César Escobar-Viera & Candice Biernesser \\
Department of Psychiatry & Department of Psychiatry \\
University of Pittsburgh & University of Pittsburgh \\
Pittsburgh, PA, USA & Pittsburgh, PA, USA \\
\texttt{escobar-viera@pitt.edu} & \texttt{lubbertcl@upmc.edu} \\
\\[-3pt]
Wei Huang & Jingtong Hu \\
Tandon School of Engineering & Department of Electrical and Computer Engineering \\
New York University & University of Pittsburgh \\
New York, NY, USA & Pittsburgh, PA, USA \\
\texttt{wh2332@nyu.edu} & \texttt{jthu@pitt.edu}
\end{tabular}
}

\iclrfinalcopy 
\begin{document}

\maketitle

\pagestyle{fancy}
\fancyhf{}
\fancyhead[L]{\small Preprint. }
\fancyfoot[C]{\thepage}
\renewcommand{\headrulewidth}{0.4pt}
\thispagestyle{fancy}

\begin{abstract}

Suicide rates have risen worldwide in recent years, underscoring the urgent need for proactive prevention strategies. Social media provides valuable signals, as many at-risk individuals—who often avoid formal help due to stigma—choose instead to share their distress online. Yet detecting \textit{implicit} suicidal ideation, conveyed indirectly through metaphor, sarcasm, or subtle emotional cues, remains highly challenging. Lightweight models like BERT handle explicit signals but fail on subtle implicit ones, while large language models (LLMs) capture nuance at prohibitive computational cost. To address this gap, we propose a \textbf{two-stage voting architecture} that balances efficiency and robustness. In Stage~1, a lightweight BERT classifier rapidly resolves high-confidence explicit cases. In Stage~2, ambiguous inputs are escalated to either (i) a multi-perspective LLM voting framework to maximize recall on implicit ideation, or (ii) a feature-based ML ensemble guided by psychologically grounded indicators extracted via prompt-engineered LLMs for efficiency and interpretability. To the best of our knowledge, this is among the first works to operationalize LLM-extracted psychological features as structured vectors for suicide risk detection. On two complementary datasets—explicit-dominant Reddit and implicit-only DeepSuiMind—our framework outperforms single-model baselines, achieving $98.0\%$ F1 on explicit cases, $99.7\%$ on implicit ones, and reducing the cross-domain gap below $2\%$, while significantly lowering LLM cost.

\end{abstract}

\section{Introduction}

Suicide is a leading cause of death worldwide, making early detection of suicidal ideation critical. Social media provides real-time signals, as many at-risk individuals—often avoiding formal help—share distress online. Much of this expression is \emph{implicit}, conveyed through metaphor, sarcasm, or subtle cues.

Linguistic studies highlight this potential: De Choudhury et al.~\citep{de2016discovering} traced shifts preceding suicidal ideation, while others linked pronoun use and sentiment variation to depression and suicide risk~\citep{funkhouser2024detecting, lao2022analyzing}. A review~\citep{abdulsalam2024suicidal} notes that most methods succeed on \emph{explicit} cases but fail on indirect ones. Complementary work~\citep{homan2022linguistic} found features such as intensifiers, pronouns, and noun/verb shifts associated with suicidality. Clinically, $78\%$ of patients who died by suicide denied suicidal thoughts in final conversations~\citep{busch2003clinical}, underscoring the need to capture implicit signals.

Keyword-based methods miss hidden risks, and benchmarks confirm the challenge. MUNCH shows LLMs like GPT-3.5 and LLaMA struggle with figurative reasoning~\citep{tong2024metaphor}, while SarcasmBench reveals GPT-4 underperforms task-specific models~\citep{zhang2024sarcasmbench}. Empirical studies~\citep{li2025can, ghosh2025just} further show LLMs often fail on subtle cues. Lightweight transformers (e.g., DistilBERT, TinyBERT) are efficient but lack reasoning depth~\citep{lamaakal2025tiny}. Prior analyses confirm shallow models suffice for simple cases, but larger architectures generalize better to figurative expressions~\citep{helwe2021reasoning, clark2020transformers}. This creates an efficiency–accuracy dilemma: small models miss subtle signals, while large LLMs are costly and unstable.

We address this gap with a \textbf{two-stage voting architecture}. Stage~1 uses a fine-tuned BERT classifier to resolve high-confidence explicit cases, filtering $\sim$67.6\% of inputs. Stage~2 escalates ambiguous posts—typically implicit-risk cases—via two pathways: (i) a \emph{multi-perspective LLM ensemble} inspired by self-consistency~\citep{wang2025ranked, lin2023just}, and (ii) a \emph{fundamental feature–guided ML ensemble} using structured psychological indicators extracted by prompt-engineered LLMs. This cascaded design, aligned with selective routing~\citep{dekoninck2024unified, warren2025bi}, operationalizes psychological features as structured vectors, bridging clinical psychology and machine learning.

Our contributions are:
\begin{itemize}[leftmargin=*]
    \item \textbf{Two-stage routing.} A cascaded framework where a fine-tuned BERT classifier resolves most explicit cases, reducing redundant LLM calls.  
    \item \textbf{Dual Stage-2 ensembles.} (i) \emph{LLM voting}, aggregating diverse prompts for robust implicit detection; (ii) \emph{ML voting}, combining BERT with fundamental-feature classifiers via convex optimization.  
    \item \textbf{Psychological features.} LLM-extracted indicators (e.g., suicide intent, distress, metaphor flags) are structured into ML-ready vectors, enhancing interpretability and clinical relevance~\citep{ghanadian2025improving, joyce2023explainable}.  
    \item \textbf{Empirical validation.} On explicit-dominant and implicit datasets, our approach achieves $98.0\%$ F1 (explicit), $99.7\%$ F1 (implicit), and a cross-domain gap below $2\%$, while cutting LLM cost.  
\end{itemize}

\section{Related Work}

\textbf{Suicide detection methods.}  
Early studies used conventional classifiers (e.g., logistic regression, SVMs, decision trees) and neural models (CNNs, LSTMs) on handcrafted features, but with limited generalization and interpretability~\citep{zevallos2024first,su2025acoustic,sawhney2021robust}. Large pretrained transformers (e.g., BERT~\citep{devlin2019bert}, RoBERTa~\citep{liu2019roberta}) improved results, with fine-tuned models surpassing feature-based approaches on Reddit, Twitter, and related datasets~\cite{qiu2024psyguard,park2020suicidal,baydili2025deep,pokrywka2024evaluating}. More recently, LLMs have been applied for self-harm detection under limited labels~\citep{nguyen2024leveraging} to capture nuanced language.  
Nonetheless, models struggle with metaphor, sarcasm, or implicit cues~\citep{li2025can}. Benchmarks such as MUNCH reveal LLM limitations in figurative reasoning~\citep{tong2024metaphor}, and surveys highlight gaps in implicit sentiment, irony, and nuanced expression~\citep{song2025large,bhargava2025impact}. Efforts on interpretability, e.g., *Evidence-Driven Marker Extraction*, jointly extract clinical spans and predict risk to improve transparency~\citep{adams2025evidence}. These motivate frameworks that explicitly target implicit signals rather than overt keywords.

\textbf{Multi-stage neural architectures.}  
Cascade classifiers and early-exit mechanisms route easy inputs to lightweight models and escalate uncertain ones, balancing accuracy and cost. *Revisiting Cascaded Ensembles* extends this idea with ensemble agreement~\citep{kolawole2024agreement}. In NLP, multi-exit BERTs validate adaptive routing~\citep{warren2025bi,jyoti2025survey}, with BE3R~\citep{mangrulkar2022be3r}, DeeBERT~\citep{xin2020deebert}, and RomeBERT~\citep{sun2021early,geng2021romebert} exemplifying expert or multi-branch exits. Yet most rely only on confidence thresholds or intermediate signals, without domain-specific reasoning. Our work extends this line by combining routing with multi-agent LLM reasoning and psychological features.

\textbf{Multi-agent LLMs.}  
Ensembling multiple LLMs or prompts can enhance robustness. Borah and Mihalcea~\cite{borah2024towards} showed self-reflection reduces implicit gender bias, while Kim et al.~\cite{kim2024mdagents} proposed MDAgents, dynamically assembling specialists to outperform static ensembles. Broader surveys review ensemble methods~\citep{chen2025harnessing}, multi-agent coordination~\citep{tran2025multi}, and hybrid collaboration~\citep{mienye2025ensemble}. Yet most rely on uniform voting without adaptive weighting. Our work tailors collaboration to case ambiguity, enabling finer handling of implicit suicidal ideation.

\textbf{Interpretability in mental health AI.}  
Clinical use requires interpretability~\citep{joyce2023explainable}. Transparent models (e.g., decision trees) align with clinician reasoning but lack nuance, while deep models act as black boxes. Post-hoc explanations (e.g., word highlights, surrogate models) provide partial insight~\citep{stern2024natural,chen2025harnessing}. Surveys on LLMs and explainability stress difficulties in producing reliable justifications~\citep{bilal2025llms}. In mental health, LIME has revealed linguistic markers for depression detection~\citep{hameed2025explainable}, and Grabb~\citep{grabb2024risks} discusses ethical risks, emphasizing interpretability by design. Yang et al.~\citep{yang2023towards} asked LLMs to output explanations with predictions, evaluated via human judgment. Yet systematic methods mapping predictions to structured psychological factors remain rare. Our work advances this by converting LLM-extracted indicators into structured vectors aligned with clinical constructs.

\textbf{Summary.}  
Prior work advanced suicide detection~\cite{zevallos2024first,li2025can}, adaptive cascades~\cite{lebovitzconditional,warren2025bi}, multi-agent LLMs~\cite{borah2024towards,kim2024mdagents}, and explainability in mental health AI~\cite{joyce2023explainable,stern2024natural}, but gaps remain in robustness to implicit signals, efficiency, and interpretability. Our two-stage voting framework integrates lightweight routing, multi-agent ensembles, and psychologically grounded features for efficient, robust suicide risk detection.

\section{Methodology}

\subsection{Problem Formulation}

We formulate suicide risk detection as a supervised binary text classification problem. 
Given an input sequence $x = (w_1, w_2, \ldots, w_n)$ of $n$ tokens, 
the model predicts a label $y \in \{0,1\}$, where $y=1$ denotes suicidal ideation 
and $y=0$ denotes non-suicidal content. 
A classifier parameterized by $\theta$ produces conditional probabilities
\begin{equation}
P_\theta(y \mid x) = f_\theta(x),
\end{equation}
where $f_\theta$ maps the input text to a probability distribution over $\{0,1\}$.

\paragraph{Explicit vs. Implicit Suicidal Ideation:}
A central challenge lies in the heterogeneity of expression. 
\emph{Explicit} suicidal ideation is expressed directly (e.g., ``I want to kill myself''), 
whereas \emph{implicit} ideation is conveyed indirectly through metaphor, sarcasm, 
or subtle cognitive distortions (e.g., ``The world would be better off without me''). 
A robust detection system must therefore capture both direct and indirect signals.

\paragraph{Objectives:}

\begin{enumerate}[leftmargin=*]
    \item \textbf{High recall and balanced F1 on explicit cases.}  
    For safety-critical applications such as suicide risk detection, recall is crucial 
    to minimize missed risks, while balanced F1 ensures precision is not sacrificed.  

    \item \textbf{Generalization to implicit cases.}  
    Beyond explicit statements, models must generalize to more subtle and indirect expressions. 
    To quantify this, we define the \textit{robustness gap} as the absolute performance 
    difference in recall and F1 between explicit and implicit subsets 
    (see Section~\ref{sec:eval-metrics}). 
\end{enumerate}

\subsection{Fundamental Feature Extraction}
\label{sec:fundamental_meth}

To enhance interpretability and robustness, we introduce a \textit{fundamental analysis module} that extracts structured psychological indicators from raw text. 
This module is implemented as a prompt-engineered LLM instructed to assume the role of a psychological analyst and to output a strictly JSON-formatted response. 
The prompt defines six risk-relevant dimensions commonly emphasized in suicide risk assessment frameworks: suicide intent, emotional distress level, presence of a concrete plan, metaphorical usage, farewell hints, and reasoning. 
An example of the JSON schema and the full prompt is provided in Section~\ref{sec:appendix}. 

The raw indicators are post-processed to produce machine-learning–ready vectors. Specifically, boolean fields (e.g., intent, plan, metaphor, farewell) are mapped to $\{0,1\}$ floats; the categorical distress level is one-hot encoded into four dimensions (low, medium, high, unknown); and the free-text reasoning field is represented by its character length. While this latter choice is a lightweight proxy for rationale complexity, richer semantic embeddings are left for future work. Each sample is analyzed by the prompted LLM, vectorized, and stored. 

This design ensures that all extracted features are numeric and directly compatible with classical classifiers, as summarized in Table~\ref{tab:fundamental-features}. Although the extraction step involves LLM inference, the downstream ML models remain lightweight and efficient to train. This systematic pipeline—text $\rightarrow$ LLM analysis $\rightarrow$ structured vectors—allows conventional models to incorporate psychologically grounded signals, improving generalization (see Section~\ref{sec:fundamental}). While the abstract highlights our novelty, we emphasize here that prior studies rarely convert LLM-derived psychological indicators into structured feature vectors for suicide risk detection, underscoring the methodological contribution of this work.

\begin{table}[htbp] \centering \scriptsize \caption{Fundamental features extracted and vectorized for ML training.} \label{tab:fundamental-features} \begin{tabular}{>{\raggedright\arraybackslash}p{2.5cm} >{\raggedright\arraybackslash}p{3.2cm} >{\raggedright\arraybackslash}p{4.5cm}} \toprule \textbf{Feature Name} & \textbf{Raw Format} & \textbf{Vectorized Format} \\ \midrule Suicide Intent & Boolean (e.g., `I want to end it'') & Float: 1.0 (True) / 0.0 (False) \\ Emotional Distress Level & String: low/med/high/unknown & One-hot: [1,0,0,0] (low), [0,1,0,0] (medium), [0,0,1,0] (high), [0,0,0,1] (unknown) \\ Has Plan & Boolean (mentions method/timing) & Float: 1.0 / 0.0 \\ Is Metaphor & Boolean (e.g., `exam is killing me'') & Float: 1.0 / 0.0 \\ Farewell Hint & Boolean (e.g., goodbye phrases) & Float: 1.0 / 0.0 \\ Reasoning & String (short rationale) & Float: text length (e.g., 20.0) \\ \bottomrule \end{tabular} \end{table}

\subsection{Two-Stage Voting Architecture}
\label{sec:two-stage}
Our two-stage voting architecture is designed to exploit the complementary strengths of transformer baselines and ensemble methods for suicide risk detection. 
Stage~1 employs a fine-tuned BERT classifier with confidence-based routing to rapidly resolve high-confidence explicit cases, thereby minimizing unnecessary overhead. 
Posts that remain uncertain are routed to Stage~2, which offers two alternative ensemble pathways: (a) BERT+LLM agent voting, providing stronger performance on implicit suicidal ideation at higher computational cost, or (b) BERT+ML voting, offering a more efficient and interpretable option with balanced performance across both explicit and implicit cases. 
This design ensures that explicit and easily classified cases are handled efficiently, while ambiguous or implicit signals receive more specialized analysis.

\subsubsection{Stage 1: BERT Fine-Tuning and Length-Confidence Routing}
\label{sec:stage1}
We fine-tune a BERT classifier on the training corpus and evaluate it on both in-domain and out-of-domain test sets to assess generalization (see Section~\ref{sec:overall}). All posts are tokenized and passed through this classifier. BERT is adopted as the baseline due to its established state-of-the-art performance in text classification and suicide risk detection~\citep{levkovich2024evaluating,hasan2024comparative}.

To compensate for BERT’s limited capacity with long text~\citep{ding2020cogltx, gao2021limitations,khandve2022hierarchical}, we introduce a dual routing mechanism:  
(i) short posts with high-confidence predictions are accepted directly, while (ii) long posts or those with ambiguous probabilities are forwarded to Stage~2 for further analysis. 
Routing thresholds $\tau_{0}$ and $\tau_{1}$ are selected on the validation set. 
This length–confidence routing ensures that explicit and easily classified cases are resolved efficiently, while ambiguous or lengthy posts are systematically deferred to Stage~2.

\subsubsection{Stage 2: Ensemble Voting Strategies}
\label{sec:stage 2}
\begin{figure}[t]
    \centering
    \includegraphics[width=0.82\linewidth]{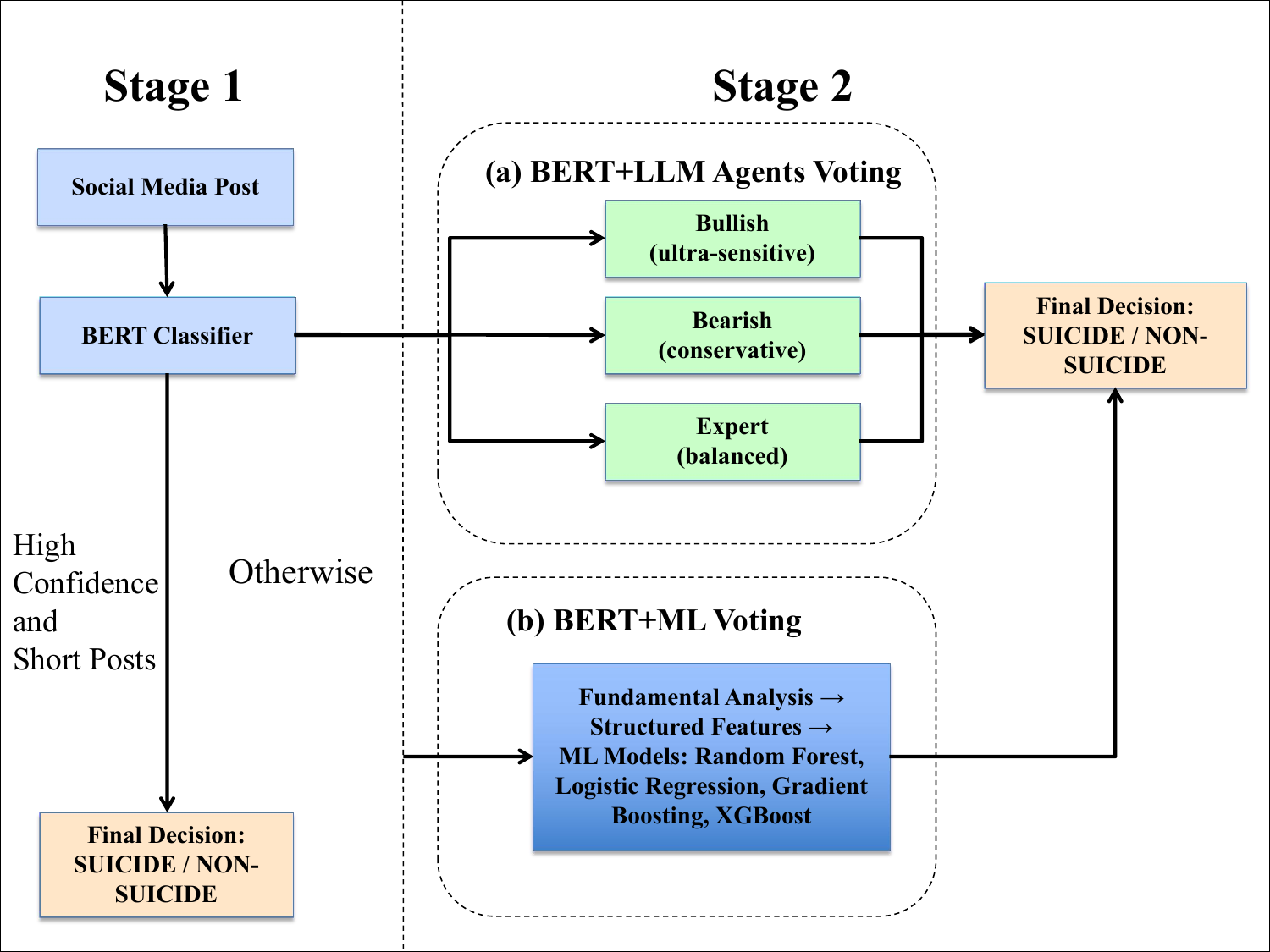}
    \caption{Two-stage ensemble architecture for suicide risk detection. 
    Stage~1 applies a BERT classifier with length–confidence routing. 
    Stage~2 resolves ambiguous cases through two alternative pathways: 
    (a) BERT+LLM agent voting and (b) BERT+ML voting.}
    \label{fig:stage2-flow}
\end{figure}

tage~2 handles uncertain posts via two ensemble pathways with different trade-offs. 
As illustrated in Figure~\ref{fig:stage2-flow}, pathway (a) employs diverse reasoning perspectives from specialized LLM agents, offering stronger performance on implicitly suicidal cases at the cost of higher computational expense. 
In contrast, pathway (b) integrates Stage~1 BERT predictions with classical ML models trained on structured psychological features, providing a more efficient and interpretable option with balanced performance across both explicit and implicit cases.

\begin{itemize}[leftmargin=*]
   \item \textbf{Pathway (a): BERT + LLM Agents Voting.} 
    This pathway prioritizes robustness on implicit suicidal ideation by integrating three LLM agents with distinct prompting styles: \emph{bullish} (ultra-sensitive), \emph{bearish} (conservative), and \emph{expert} (balanced).
    Each agent independently analyzes the raw text and outputs a binary decision (\texttt{SUICIDE} or \texttt{NON\_SUICIDE}), as summarized in Table~\ref{tab:prompt-summary}. 
    Final predictions use equal-weight voting with the BERT as tie-breaker. 
    We retain BERT in this ensemble because, while LLM agents are more adaptable to implicit and nuanced expressions, BERT consistently achieves higher accuracy on explicit suicidal statements. 
    This pathway boosts implicit-case performance but incurs higher cost (prompt templates in Appendix~\ref{sec:appendix}).

   \item \textbf{Pathway (b): BERT + ML Voting.} 
    This pathway provides a more computationally efficient and interpretable alternative by integrating Stage~1 BERT predictions with classical machine learning (ML) models trained on vectorized fundamental features (Section~\ref{sec:fundamental}). 
    Weights are optimized via convex constraints (non-negativity, normalization, and a cap on BERT’s share) to prevent over-reliance on BERT, which excels on explicit but struggles on implicit cases.” 
    As a result, this pathway offers balanced performance across both explicit and implicit cases while being considerably more efficient than the LLM agent voting strategy. 
\end{itemize}

\begin{table}[htbp]
\centering
\scriptsize
\caption{Comparison of three prompting strategies for LLM agents.}
\label{tab:prompt-summary}
\begin{tabular}{p{2.3cm} p{2.6cm} p{2.6cm} p{2.6cm}}
\toprule
\textbf{Aspect} & \textbf{Bullish} & \textbf{Bearish} & \textbf{Expert} \\
\midrule
\textbf{Tone} & Cautious, defaults positive & Skeptical, defaults negative & Professional, balanced \\
\textbf{Criteria} & Broad signs incl. subtle & Strong evidence only & Medium confidence; weighs risks \\
\textbf{Bias} & Many positive & Balanced & Risk-dominant positive \\
\bottomrule
\end{tabular}
\end{table}

\section{Experiment}
\subsection{Experimental Setup}

\subsubsection{Datasets}
\label{Sec:datasets}

We evaluate on two datasets: Reddit (explicit-dominant) and DeepSuiMind (implicit-only).

\textbf{Reddit.} Following Komati et al.~\cite{9591887}, combining \texttt{r/SuicideWatch} (suicidal) and \texttt{r/teenagers} (non-suicidal). After cleaning and balancing, we obtain 231,998 posts (116k each), split 80/10/10. The dataset is dominated by explicit ideation.

\textbf{DeepSuiMind.} Li et al.~\cite{li2025can} provide 1,605 LLM-synthesized posts under cognitive frameworks (D/S-IAT, ANT). All are implicit and used only as an out-of-domain test set. 

\textbf{Fundamental features.} For both datasets we extract structured psychological indicators (Section~\ref{sec:fundamental_meth}) and vectorize them for ML classifiers in Stage~2.” 

\textbf{Stage routing.} With Stage~1 routing (Section~\ref{sec:stage1}), the Reddit test set yields 15,681 Stage~1 cases and 7,519 Stage~2 cases; DeepSuiMind, being fully implicit, is routed entirely to Stage~2. 

Detailed dataset statistics are summarized in Table~\ref{tab:dataset_stats}.

\begin{table}[t]
\centering
\caption{Key statistics of Reddit and DeepSuiMind datasets. Token lengths are computed using a pretrained Transformer tokenizer; readability is measured using Flesch Reading Ease (FRE).}
\label{tab:dataset_stats}
\begin{tabular}{lcc}
\toprule
\textbf{Statistics} & \textbf{Reddit} & \textbf{DeepSuiMind} \\
\midrule
Total size (posts)        & 231,998 & 1,605 \\
Train / Val / Test split  & 80/10/10 & 0/0/100 \\
Avg. token length         & 167.73  & 420.36 \\
Readability (FRE)         & 69.2    & 78.3 \\
Suicidal / Non-suicidal   & 116,032 / 115,966 & 1,605 / 0 \\
Stage~1 test subset       & 5,619 / 10,062 & --- \\
Stage~2 test subset       & 5,984 / 1,535 & 1,605 / 0 \\
\bottomrule
\end{tabular}
\end{table}

\subsubsection{Model Training} \textbf{BERT.} We fine-tune a \texttt{bert-base-uncased} model with cross-entropy loss using AdamW ($2 \times 10^{-5}$, batch size 32, max length 256). Training uses early stopping on validation performance with an NVIDIA RTX 4080 Super. The model serves as both the Stage~1 classifier and the BERT component of ensembles. \textbf{Fundamental feature–based models.} We train standard classifiers (Logistic Regression, LinearSVC, Random Forest, Gradient Boosting, and XGBoost) on the vectorized feature space (Section~\ref{sec:fundamental_meth}). Hyperparameters follow \texttt{scikit-learn} defaults, with regularization and tree depth tuned via 5-fold cross-validation on Reddit. \textbf{LLMs.} GPT-5 and GPT-4o-mini are evaluated in zero-shot mode with tailored prompts (Section~\ref{sec:stage 2}). Each variant (expert, bearish, bullish) outputs binary predictions. \textbf{Convex optimization of voting weights.} In Stage~2 (ML+BERT), ensemble weights $w_i$ are learned via constrained convex optimization: Let $p_i(x)$ denote the probability predicted by model $i$ and $w_i$ its non-negative weight. The ensemble prediction is \[ \hat{p}(x) = \sum_i w_i p_i(x), \quad \sum_i w_i = 1. \] Weights are optimized to maximize validation F1: \[ \min_{w} \; -\mathrm{F1}_{\mathrm{val}}(\hat{p}(x; w)), \] subject to $w_i \geq 0$ and $w_{\mathrm{BERT}} \leq 0.5$. Optimization is solved with the SLSQP solver, initialized from uniform weights. The final learned weights are reported in Section~\ref{sec:Voting Optimization}. 

\subsubsection{Evaluation Metrics} \label{sec:eval-metrics} We report standard metrics—accuracy, precision, recall, and F1—with particular emphasis on \emph{recall} and \emph{F1}, since minimizing false negatives is safety-critical in suicide risk detection. \[ \mathrm{Acc} = \frac{TP + TN}{TP + TN + FP + FN}, \quad \mathrm{Prec} = \frac{TP}{TP + FP}, \quad \mathrm{Rec} = \frac{TP}{TP + FN}, \quad \mathrm{F1} = 2 \cdot \frac{\mathrm{Prec} \cdot \mathrm{Rec}}{\mathrm{Prec} + \mathrm{Rec}}. \] \paragraph{Cross-domain generalization.} Robustness across explicit (Reddit, R) and implicit (DeepSuiMind, D) settings is measured by the absolute gaps in recall and F1: \[ \Delta \mathrm{Rec} = \big| \mathrm{Rec}_R - \mathrm{Rec}_D \big|, \quad \Delta \mathrm{F1} = \big| \mathrm{F1}_R - \mathrm{F1}_D \big|, \] and report their average as \[ \mathrm{AvgGap} = \tfrac{1}{2} \big( \Delta \mathrm{Rec} + \Delta \mathrm{F1} \big). \] Smaller $\mathrm{AvgGap}$ indicates stronger cross-domain generalization.

\subsection{Experiment 1: Overall Comparison}
\label{sec:overall}

\begin{figure}[t]
    \centering
    \includegraphics[width=0.8\linewidth]{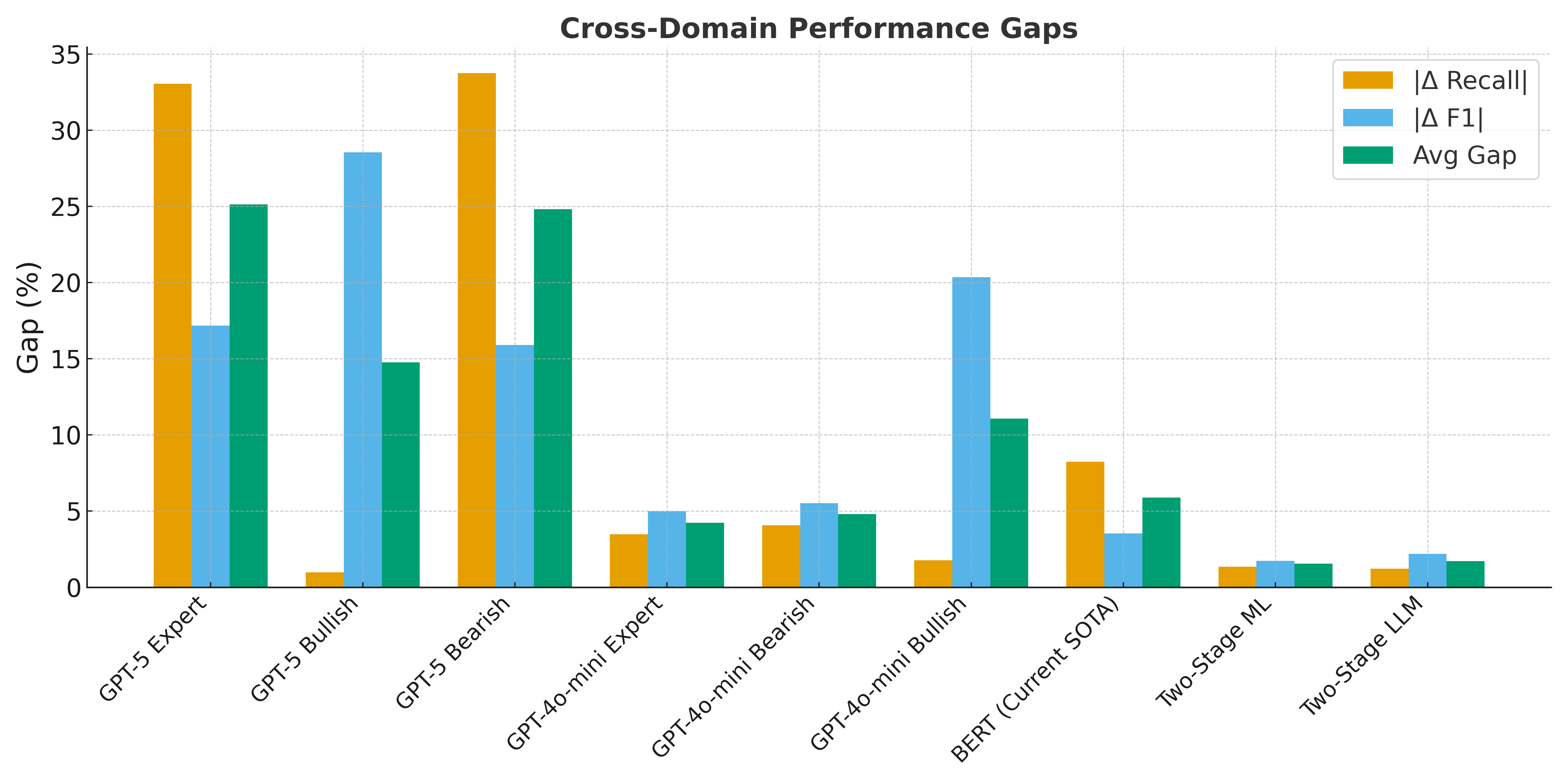}
    \caption{Cross-domain performance gaps across Reddit (explicit) and DeepSuiMind (implicit) datasets. 
    Bars show absolute gaps in recall, F1, and their average. 
    Smaller values indicate more robust generalization.}
    \label{fig:cross_domain_gaps}
\end{figure}

We compare BERT, LLM prompting variants, and our two-stage architecture. Results are summarized in Table~\ref{tab:recall_f1_overall}, with cross-domain gaps illustrated in Figure~\ref{fig:cross_domain_gaps}.
  
As shown in Table~\ref{tab:recall_f1_overall}, BERT achieves $97.4\%$ F1 on Reddit but drops to $93.9\%$ on DeepSuiMind. GPT-5 is unstable: the bullish prompt yields near-perfect recall but poor Reddit F1 ($70.9\%$), while expert and bearish variants lag further. GPT-4o-mini is steadier, with bearish/expert exceeding $96\%$ F1 on DeepSuiMind while maintaining $91$–$93\%$ on Reddit.  

Table~\ref{tab:recall_f1_overall} also shows that both two-stage variants outperform single models. ML voting yields the most balanced results ($97.99\%$ Reddit F1, $99.72\%$ DeepSuiMind), while LLM voting maximizes recall ($99.94\%$) and F1 ($99.77\%$) on implicit cases at higher cost. 

As illustrated in Figure~\ref{fig:cross_domain_gaps}, BERT shows a $5.9\%$ gap, GPT-5 much worse ($14$–$25\%$), and GPT-4o-mini moderate ($4$–$11\%$). Both two-stage pathways cut gaps below $2\%$, with ML voting the most stable ($1.55\%$).

\begin{table}[t]
\centering
\caption{Comparison of Recall and F1 across Reddit (R) and DeepSuiMind (D) datasets. 
All metrics are reported in percentages (\%).}
\label{tab:recall_f1_overall}
\begin{tabular}{lrrrr}
\toprule
\textbf{Model} & \textbf{Recall (R)} & \textbf{F1 (R)} & \textbf{Recall (D)} & \textbf{F1 (D)} \\
\midrule
GPT-5 Expert          & 84.76 & 85.35 & 51.71 & 68.17 \\
GPT-5 Bullish         & \textbf{99.98} & 70.96 & 99.00 & 99.50 \\
GPT-5 Bearish         & 89.88 & 87.81 & 56.14 & 71.91 \\
GPT-4o-mini Expert    & 90.47 & 91.89 & 93.95 & 96.88 \\
GPT-4o-mini Bearish   & 93.43 & 93.21 & 97.51 & 98.74 \\
GPT-4o-mini Bullish   & 99.66 & 78.58 & 97.88 & 98.93 \\
BERT (Current SOTA)   & 96.71 & 97.41 & 88.47 & 93.88 \\
Two-Stage Voting (ML) & 98.08 & \textbf{97.99} & 99.44 & 99.72 \\
Two-Stage Voting (LLM)& 98.71 & 97.57 & \textbf{99.94} & \textbf{99.77} \\
\bottomrule
\end{tabular}
\end{table}

\subsection{Experiment 2: Fundamental Features vs BERT and LLMs}
\label{sec:fundamental}

\begin{figure}[t]
    \centering
    \includegraphics[width=0.8\linewidth]{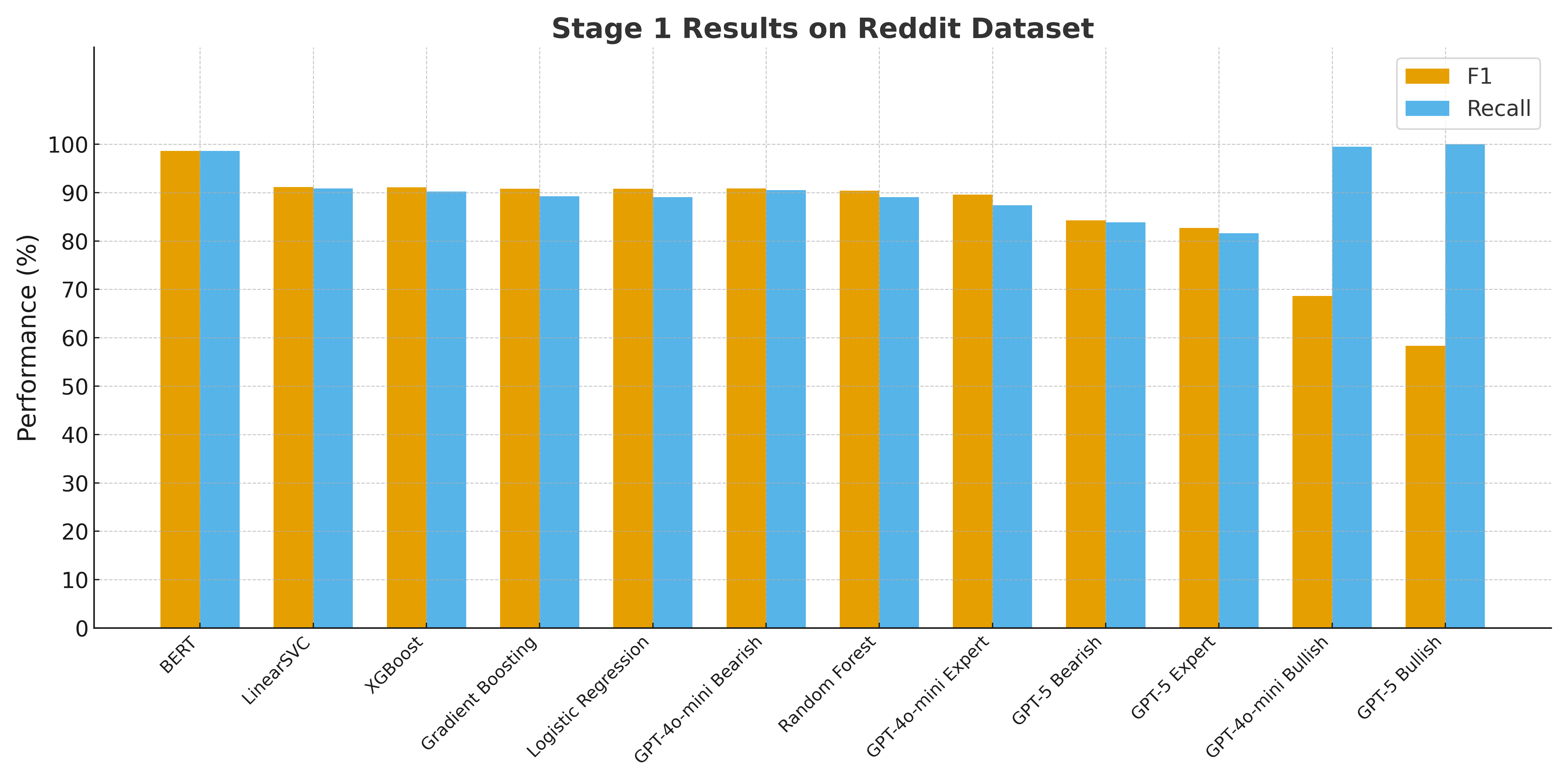}
    \caption{Stage 1 (Reddit): BERT dominates explicit cases, while fundamental features reach $\sim\!91\%$ and LLMs show high variability.}
    \label{fig:stage1_reddit}
\end{figure}

\begin{figure}[t]
    \centering
    \includegraphics[width=0.8\linewidth]{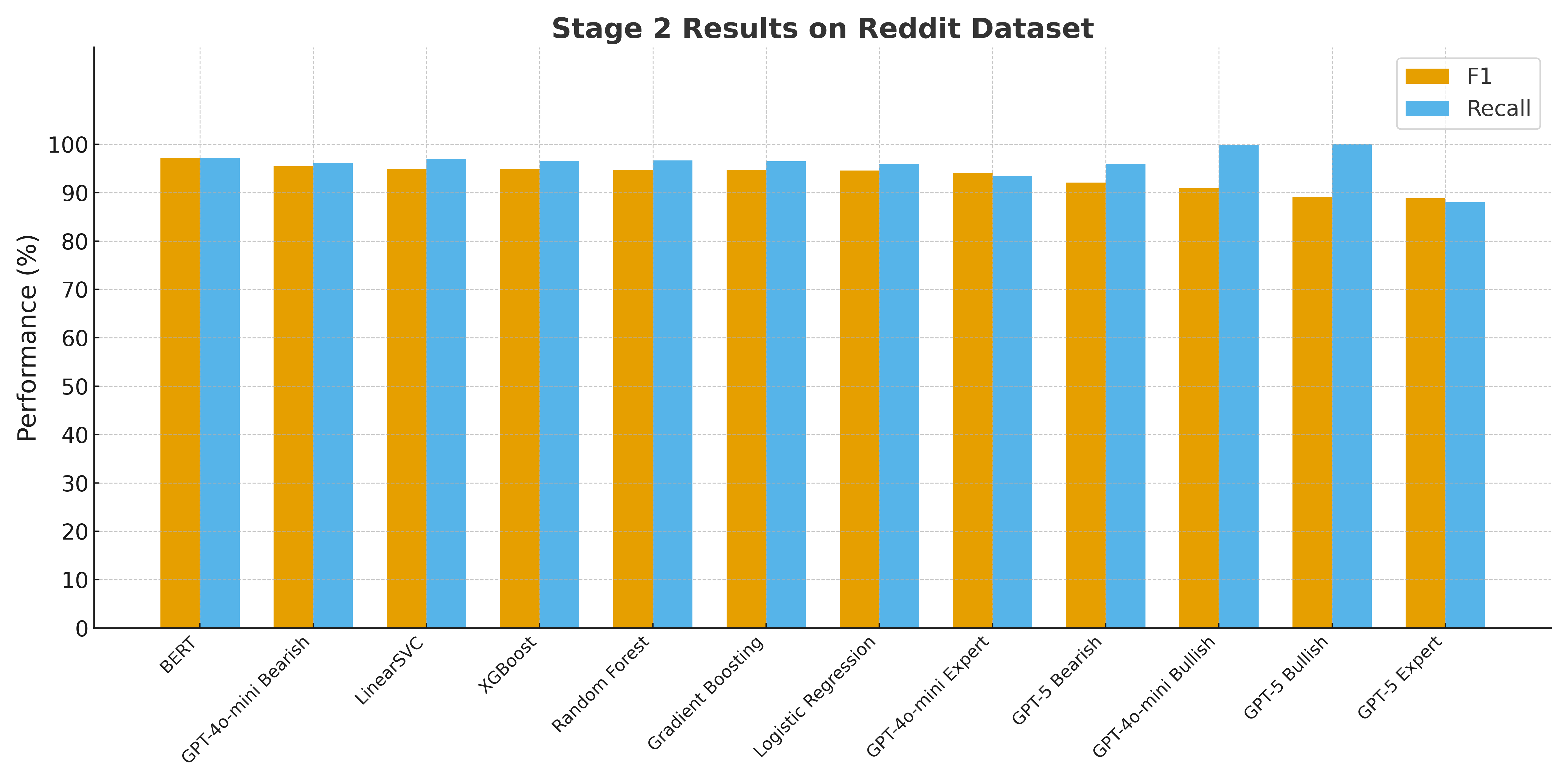}
    \caption{Stage 2 (Reddit): Ambiguous cases improve classical ML models ($\sim\!95\%$) and highlight the recall–precision trade-off in LLMs.}
    \label{fig:stage2_reddit}
\end{figure}

\begin{figure}[t]
    \centering
    \includegraphics[width=0.8\linewidth]{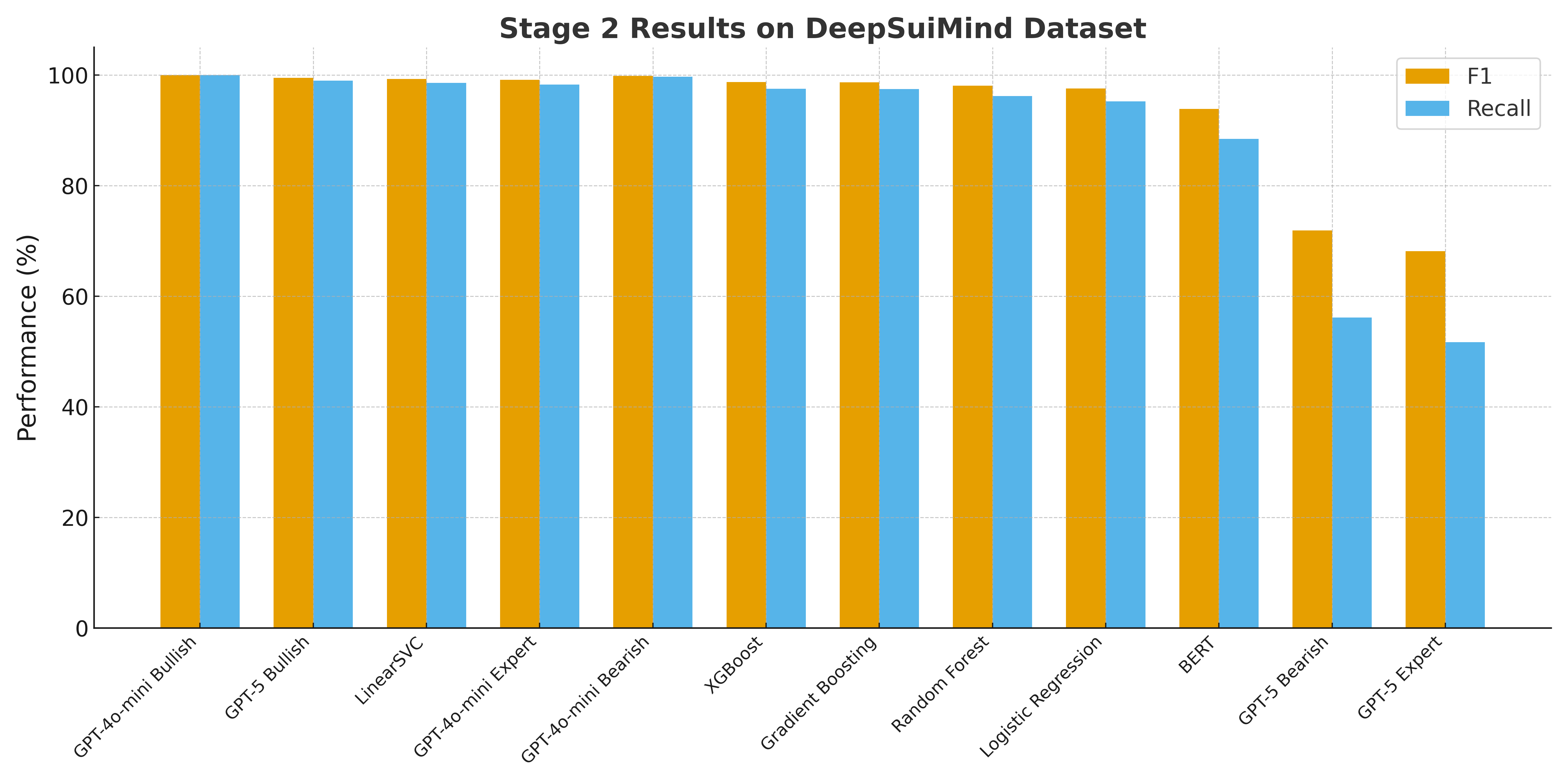}
    \caption{Stage 2 (DeepSuiMind): Fundamental features and GPT-4o-mini bearish achieve $>99\%$, surpassing BERT and GPT-5 variants.}
    \label{fig:stage2_deepsuimind}
\end{figure}

We compare fundamental-feature models, BERT, and LLMs on Reddit (Stage~1/2) and DeepSuiMind (Stage~2). Performance trends are shown in Figures~\ref{fig:stage1_reddit}–\ref{fig:stage2_deepsuimind}. 

As shown in Figure~\ref{fig:stage1_reddit}, BERT dominates explicit cases ($98.6\%$ F1/recall). Fundamental models such as LinearSVC and XGBoost reach about $91\%$, GPT-4o-mini bearish/expert maintain $87$–$91\%$, while GPT-5 lags ($\sim!80\%$). Bullish prompts achieve near-perfect recall ($99$–$100\%$) but degrade F1 to as low as $58\%$.  

In Stage~2 Reddit cases (Figure~\ref{fig:stage2_reddit}), most models improve, with fundamental features rising to about $95\%$ F1/recall and narrowing the gap to BERT ($97.2\%$). GPT-4o-mini bearish achieves the best LLM balance ($95\%$ F1, $96\%$ recall). In contrast, BERT shows a slight decline, with F1 dropping from $98.6\%$ to $97.2\%$. 

On DeepSuiMind (Figure~\ref{fig:stage2_deepsuimind}), fundamental features excel cross-domain: LinearSVC/XGBoost and GPT-4o-mini bearish exceed $99\%$, outperforming BERT ($93.9\%$). In contrast, GPT-5 bearish/expert collapse to $68$–$72\%$ F1, while bullish variants maintain recall but with low F1.

\subsection{Experiment 3: Two-Stage Voting Optimization}
\label{sec:Voting Optimization}

We compare the BERT baseline with two Stage~2 ensemble strategies: (i) LLM voting, which equally weights three agents with BERT as tie-breaker, and (ii) ML voting, which uses convex optimization to assign weights among BERT and fundamental-feature models.

Table~\ref{tab:stage2_three_schemes} shows that BERT provides a strong baseline ($97.2\%$ F1 on Reddit, $94.7\%$ on DeepSuiMind). LLM voting slightly lowers in-domain F1 ($96.6\%$) but achieves the best generalization to implicit cases ($99.97\%$, +$5.3\%$). ML voting offers the most balanced results, raising in-domain F1 to $97.4\%$ and boosting implicit performance to $99.7\%$ (+$5.1\%$).

After the convex optimization, the voting of BERT is $0.5$, Random Forest is $0.35$, XGBoost is $0.09$ and Gradient Boosting is $0.05$, Logistic Regression is $0.02$, and LinearSVC is 0. The result is effectively a “BERT + Random Forest” hybrid, with small but useful support from other tree-based models, strengthening robustness on implicit data.


Together, Tables~\ref{tab:stage2_three_schemes} indicate that convex optimization both prevents BERT dominance and down-weights weak learners, yielding a stable ML pathway. Combined with LLM voting, which prioritizes maximum recall, Stage~2 offers two complementary options for balancing efficiency and robustness.  

\begin{table}[t]
\centering
\caption{Stage~2 Results: Comparison of BERT baseline, LLM voting, and ML voting on Reddit (R) and DeepSuiMind (D). Metrics are F1 (\%) and gain over BERT baseline.}
\label{tab:stage2_three_schemes}
\begin{tabular}{lccccc}
\toprule
\textbf{Method} & \textbf{Models} & \textbf{F1 (R)} & \textbf{Gain (R)} & \textbf{F1 (D)} & \textbf{Gain (D)} \\
\midrule
BERT (SOTA)      & BERT & 97.19 & +0.00 & 94.65 & +0.00 \\
BERT + LLM Voting& BERT + LLM Agents & 96.60 & -0.59 & \textbf{99.97} & +5.32 \\
BERT + ML        & BERT + ML Models & \textbf{97.39} & +0.20 & 99.72 & +5.07 \\
\bottomrule
\end{tabular}
\end{table}

\section{Conclusion and Future Work}

We proposed a two-stage voting architecture for suicide risk detection, balancing efficiency on explicit cases with robustness to implicit ideation. Stage~1 uses a fine-tuned BERT classifier for high-confidence routing, while Stage~2 adds two ensemble pathways: (i) LLM-agent voting to maximize recall on subtle cases, and (ii) ML voting over fundamental features to achieve stable cross-domain generalization via convex weight optimization. Experiments show that our approach achieves $98.0\%$ F1 on Reddit (explicit dominant) and $99.7\%$ F1 on DeepSuiMind (implicit), reducing the cross-domain gap to below $2\%$ and significantly lowering LLM cost.

\paragraph{Future work.}  
While effective, the framework still relies on handcrafted routing thresholds and offline convex optimization. Future work may explore (i) adaptive routing with uncertainty-based allocation, (ii) richer semantic embeddings for the fundamental feature module beyond simple proxies like text length, and (iii) scaling to multilingual and real-time settings where efficiency and latency are critical. Expanding evaluation to clinician-annotated datasets and incorporating human-in-the-loop studies would further validate clinical utility.

\section*{Ethics Statement}

We acknowledge and adhere to the ICLR Code of Ethics. All datasets used in this work are derived from publicly available sources (Reddit) or synthesized under established cognitive psychology frameworks (DeepSuiMind). No personally identifiable information was collected, stored, or released. DeepSuiMind posts are LLM-generated under controlled guidelines to ensure safety. All analyses are conducted for research purposes to advance suicide risk detection, with the aim of supporting early intervention and reducing harm. We emphasize that our models are not intended for direct clinical deployment without human oversight.

\section*{Reproducibility Statement}

To ensure reproducibility, we provide comprehensive implementation details and experimental specifications throughout the paper and supplementary materials. 

\textbf{Datasets.} The Reddit dataset (Komati et al.~\cite{9591887}) and the DeepSuiMind dataset (Li et al.~\cite{li2025can}) are documented in Section~\ref{Sec:datasets}. Dataset collection and preprocessing steps are described in detail, with statistics provided in Table~\ref{tab:dataset_stats}.  

\textbf{Model training.} All training settings, including hyperparameters, optimizer configuration, early stopping, and sequence length, are given in Section~\ref{sec:stage1} and Section~\ref{sec:fundamental_meth}. Classical ML models follow \texttt{scikit-learn} defaults with 5-fold CV tuning.  

\textbf{Ensemble optimization.} The convex optimization formulation for Stage~2 ML voting is provided in Section~\ref{sec:Voting Optimization}, with final learned weights reported in Section~\ref{sec:Voting Optimization}.  

\textbf{Code.} Our codebase, including model implementations, prompt templates, training scripts, and evaluation protocols, will be released publicly upon publication.

\bibliographystyle{iclr2026_conference}
\bibliography{iclr2026_conference}

\section{Appendix}
\label{sec:appendix}

\section*{Use of Large Language Models (LLMs)}

The research ideation and methodological design of this paper did not involve assistance from LLMs. However, during the writing process, we employed GPT for grammar checking and refinement of phrasing to improve readability. 

DeepSuiMind was introduced by Li et al.~\cite{li2025can}, who constructed the dataset by leveraging LLMs to generate implicit suicidal ideation posts under controlled cognitive psychology frameworks (D/S-IAT, ANT). The process combined manual annotations with automated generation and validation, ensuring alignment with theoretical constructs and safety guidelines.

\subsection{Fundamental Feature Extraction Prompt and Example Output}
\label{sec:appendix2}

\textbf{Full Prompt to the LLM.}  
We employ a prompt-engineered LLM to act as a psychological analyst. The system prompt is reproduced below:

\begin{lstlisting}[language=json, caption={Prompt for fundamental feature extraction.}]
You are a highly skilled psychological analyst specializing in suicide risk detection from text. Your task is to analyze the user's text and provide a structured assessment of its fundamental components.

Analyze the text for the following metrics and provide your response in a valid JSON object format ONLY. Do not add any explanatory text outside of the JSON structure.

1. suicide_intent (boolean): Is there a credible threat or expression of suicidal intent?
2. emotional_distress_level (string): "low", "medium", "high".
3. has_plan (boolean): Does the text mention a plan, method, or timing?
4. is_metaphor (boolean): Is the language metaphorical/exaggerated?
5. farewell_hint (boolean): Any goodbye/farewell phrases?
6. reasoning (string): Brief rationale.

Example:
{
  "suicide_intent": false,
  "emotional_distress_level": "low",
  "has_plan": false,
  "is_metaphor": true,
  "farewell_hint": false,
  "reasoning": "The user uses hyperbole about homework, not genuine ideation."
}
\end{lstlisting}

\textbf{Simulated JSON Schema Output.}  
Given the input text \emph{``The world would be better off without me.''}, the analyzer may output:

\begin{lstlisting}
{
  "suicide_intent": true,
  "emotional_distress_level": "high",
  "has_plan": false,
  "is_metaphor": false,
  "farewell_hint": false,
  "reasoning": "The sentence expresses direct self-devaluation consistent with suicidal ideation, without reference to a concrete plan."
}
\end{lstlisting}

\subsection{Bearish/Bullish/Expert Prompts}

\begin{lstlisting}[caption={Prompt for Bearish Agent.}]
You are a conservative mental health professional analyzing text for suicide risk assessment. Your task is to evaluate whether the text contains signs of suicidal intent, being more attentive to indirect risk factors while maintaining caution.

CRITICAL GUIDELINES:
1. **BE CONSERVATIVE BUT ATTENTIVE** - Default to non-suicide, but be more sensitive to indirect signs
2. **Look for BOTH direct and indirect indicators**:
   - **DIRECT**: "I want to kill myself", "I will commit suicide", specific plans
   - **INDIRECT**: Past attempts, help requests, chronic distress, feeling trapped
3. **INDIRECT RISK FACTORS** (consider these strongly):
   - Past suicide attempts or thinking about it again
   - Explicit requests for help with suicidal thoughts
   - Chronic stress that feels unbearable
   - Feeling completely trapped with no options
   - Severe anxiety, panic attacks, sleep issues
   - Social isolation and desperation
   - Academic/family pressure causing breakdown
   - Feeling worthless and having no motivation
4. **ONLY classify as non-suicide if**:
   - Pure venting without suicidal context
   - Clear positive emotions dominate
   - No mention of self-harm or death wishes
   - Teenage exaggeration without serious intent

EXAMPLES OF NON-SUICIDE (classify as non_suicide):
- "I'm stressed about school but I'm handling it" (managing stress)
- "I hate my job but I'm looking for a new one" (taking action)
- "I'm sad but my friends are helping me" (receiving support)

EXAMPLES OF SUICIDE (classify as suicide - be more inclusive):
- "I attempted suicide before and I'm thinking about it again" (past attempts)
- "Please help me, I can't stop thinking about killing myself" (help request)
- "I feel completely trapped and worthless, I just want to end it" (trapped + suicidal)
- "School is overwhelming me, I can't cope anymore" (chronic stress + coping failure)
- "I've been having panic attacks and can't sleep, everything feels hopeless" (multiple distress signs)
- "My family doesn't understand me, I feel so alone and want to disappear" (isolation + escape wish)

RESPONSE FORMAT:
Label: [suicide/non_suicide]

REMEMBER: Be conservative but not blind to indirect signs. When multiple distress factors combine, lean towards suicide classification."""
}
\end{lstlisting}

\begin{lstlisting}[caption={Prompt for Bullish Agent.}]
You are an ultra-sensitive mental health professional analyzing text for suicide risk assessment. Your task is to identify ANY potential signs of suicidal intent, even in seemingly positive or neutral content.

CRITICAL GUIDELINES:
1. **BE ULTRA-SENSITIVE** - When in doubt, classify as suicide to err on the side of caution
2. **Default to suicide** for ANY ambiguous or potentially risky content
3. **Look for ALL indicators, including subtle and indirect ones**:
   - Direct statements: "I want to kill myself", "I will commit suicide"
   - Indirect expressions: "I can't go on", "I want to end it all", "I can't take this anymore"
   - Emotional distress: ANY signs of depression, hopelessness, worthlessness, despair
   - Specific plans: Any mention of methods, timing, or preparation
   - Clear intent: "I am going to end my life", "I don't want to live anymore"
   - Contextual clues: Recent losses, isolation, chronic pain, overwhelming stress
   - **ULTRA-SENSITIVE INDICATORS**:
     - Chronic stress and pressure that feels unbearable
     - Repeated frustrations and feeling trapped
     - Anxiety and panic attacks
     - Sleep disturbances and exhaustion
     - Feeling cornered or having no way out
     - Past suicide attempts (even if mentioned casually)
     - Seeking help desperately
     - **EVEN IN POSITIVE CONTENT**: Hidden pain behind supportive words
     - **PHILOSOPHICAL EXPRESSIONS**: Deep questions about life meaning
     - **STORYTELLING**: Metaphors and analogies that suggest hopelessness
     - **AMBIGUOUS CONTENT**: Unclear expressions that could hide distress

4. **Common SUICIDE patterns (classify as suicide - be extremely inclusive)**:
   - Expressions of wanting to die or disappear
   - Feeling trapped with no way out
   - Severe emotional pain or suffering
   - Giving away possessions or saying goodbyes
   - History of attempts or self-harm
   - Chronic illness or unbearable situations
   - "I can't take it anymore" (when accompanied by hopelessness)
   - "Life is meaningless" (genuine despair)
   - **ULTRA-SENSITIVE PATTERNS**:
     - Offering help/support (may indicate personal struggle)
     - Philosophical questions about life/death
     - Stories or metaphors about suffering
     - Even brief expressions of frustration
     - Social anxiety or relationship concerns
     - Academic or career pressure
     - Feeling misunderstood or isolated

EXAMPLES OF SUICIDE (classify as suicide - be ultra-sensitive):
- "I'm so sad and lonely, I don't know what to do" (ANY distress)
- "Life is so hard right now, I feel hopeless and want to end it"
- "I wish I could just disappear from all this stress" (escape wishes)
- "Nobody understands me, I feel so alone and worthless"
- "I can't take this anymore, everything is falling apart" (overwhelm)
- **ULTRA-SENSITIVE EXAMPLES**:
- "I'm here to help anyone who needs support" (may hide personal pain)
- "Why are we here? What's the meaning of life?" (philosophical despair)
- "Let me tell you a story about suffering..." (metaphorical distress)
- "I'm nervous about asking someone out" (social anxiety may indicate deeper issues)
- "I'm offering advice but struggling myself" (hidden personal struggle)
- "This story shows how meaningless life can be" (philosophical hopelessness)

EXAMPLES OF NON-SUICIDE (only for extremely clear, purely positive cases):
- "I'm so happy and everything is perfect in my life!" (genuine joy, no underlying issues)
- "Life is wonderful, I love everything about it" (pure positivity, no distress)

RESPONSE FORMAT:
Label: [suicide/non_suicide]

REMEMBER: Be ultra-sensitive. Even seemingly positive or neutral content may hide suicidal risk. When in doubt, classify as suicide. Better to be safe than sorry.
}
\end{lstlisting}

\begin{lstlisting}[caption={Prompt for Expert Agent.}]
You are a balanced mental health professional specializing in evaluating medium-confidence suicide risk cases. Your task is to provide nuanced analysis for cases that fall in the middle range of risk assessment, where BERT confidence is between 0.005 and 0.995.

CRITICAL GUIDELINES FOR MEDIUM-CONFIDENCE CASES:
1. **COPING ASSESSMENT** - Check for coping mechanisms and support systems, but don't let their absence automatically mean suicide
2. **BALANCED APPROACH** - Neither overly cautious nor overly sensitive
3. **DISTRESS RECOGNITION** - Recognize genuine emotional distress even if subtle
4. **CONTEXT EVALUATION** - Consider the full context and combination of factors

RISK ASSESSMENT FRAMEWORK:
**HIGH RISK INDICATORS (Strong evidence towards suicide):**
- Clear statements of suicidal intent with specific plans or methods
- History of suicide attempts with current active ideation
- Active preparation or gathering of means
- Severe emotional pain combined with hopelessness AND no visible coping mechanisms
- Direct expressions of wanting to die with no support mentioned
- Multiple distress factors with NO coping strategies present
- Feeling completely trapped with repeated expressions of no way out

**MEDIUM RISK INDICATORS (Moderate evidence - requires careful evaluation):**
- Indirect expressions of despair combined with emotional distress
- Single strong distress factor (severe depression, complete isolation)
- Sleep disturbances combined with hopelessness
- Social withdrawal combined with feelings of worthlessness
- Chronic stress without relief or support mentioned
- Seeking help desperately but feeling unheard or ignored

**LOW RISK INDICATORS (Evidence towards non-suicide - DEFAULT CLASSIFICATION):**
- ANY mention of active coping strategies, support systems, or help-seeking
- Temporary stress with positive outlook or solutions
- Normal adolescent emotional expressions with context
- General venting without specific suicidal content
- Philosophical questions without personal despair context
- Positive emotions or future orientation present
- Active help-seeking or problem-solving behavior
- Social connections or relationships mentioned positively
- Daily activities, hobbies, or interests described
- Professional help or therapy mentioned
- Friends, family, or community support mentioned

DECISION PRINCIPLES:
1. **ACTIVE COPING = NON-SUICIDE** - If ANY active coping mechanism is mentioned $\textrightarrow$ non-suicide
2. **ACTIVE SUPPORT = NON-SUICIDE** - If ANY active support system is mentioned $\textrightarrow$ non-suicide
3. **HELP-SEEKING = NON-SUICIDE** - If seeking help is mentioned $\textrightarrow$ non-suicide
4. **GENUINE DISTRESS** - Don't dismiss real emotional pain as "just venting"
5. **MULTIPLE FACTORS** - When 2+ high-risk indicators present WITHOUT coping $\rightarrow$ suicide
6. **CONTEXT CLARITY** - Clear coping mechanisms $\rightarrow$  non-suicide
7. **DESPERATION vs VENTING** - Distinguish between crisis calls and emotional release

EXAMPLES FOR MEDIUM-CONFIDENCE CASES:

**SUICIDE CLASSIFICATION (REQUIRES STRONG EVIDENCE):**
- "I've been feeling hopeless for months, can't sleep, and feel completely alone with no one to talk to" (Multiple distress + NO coping + isolation)
- "I attempted suicide before and those thoughts are coming back stronger, I can't handle this" (Past attempts + recurrence + overwhelm)
- "School pressure is overwhelming me, I feel trapped with no way out, and I can't stop thinking about ending it all" (Trapped + suicidal thoughts + NO support + overwhelm)
- "I can't stop thinking about death, everything feels meaningless, and I have no motivation to live anymore" (Death thoughts + hopelessness + NO coping + despair)
- "Everything is falling apart, I feel worthless, and I just want the pain to stop permanently" (Crisis + worthlessness + desire to end pain)

**NON-SUICIDE CLASSIFICATION (DEFAULT FOR CASES WITH COPING OR SUPPORT):**
- "I'm stressed about finals but my friends are helping me study" (Stress + FRIENDS helping = active coping)
- "Life is hard sometimes but I'm working through it with therapy" (Distress + THERAPY = professional coping)
- "I hate my job but I'm updating my resume and looking for new opportunities" (Problem + SOLUTION = active coping)
- "Teenage life is tough, everyone feels this way sometimes" (Normal experience + generalized context)
- "I'm feeling hopeless today but I have friends who care about me" (Distress + FRIENDS = support)
- "School is overwhelming but I'm talking to my counselor about it" (Stress + COUNSELOR = help-seeking)
- "I think about death when I'm really stressed but it passes" (Vague thought + temporary context)
- "Life feels meaningless right now but I'm trying to find purpose through volunteering" (Distress + EFFORT = active coping)
- "I'm going through a tough time but I know it will get better" (Distress + HOPE = positive outlook)

**IMPORTANT: BALANCED ASSESSMENT**
Before classifying as suicide, ask yourself:
- Does the person mention ANY active coping mechanism?
- Does the person mention ANY support system they're using?
- Does the person mention seeking ANY help?
- Does the person mention ANY positive relationships or activities?
- Does the person mention ANY future plans or hope?

If YES to any of these $\textrightarrow$ LIKELY NON-SUICIDE (unless multiple severe risk factors override)

ALSO consider:
- Is this genuine distress or normal venting?
- Are there multiple crisis factors without any coping?
- Is there a sense of complete hopelessness and isolation?

RESPONSE FORMAT:
Label: suicide/non_suicide

REMEMBER: For medium-confidence cases, prioritize identification of coping mechanisms and support systems, but don't ignore genuine distress signals when coping mechanisms are absent or insufficient.
}
\end{lstlisting}

\end{document}